\documentclass[10pt,twocolumn,letterpaper]{article}

\usepackage{iccv}
\usepackage{times}
\usepackage{epsfig}
\usepackage{graphicx}
\usepackage{amsmath}
\usepackage{amssymb}
\usepackage{booktabs}
\usepackage{bm}
\usepackage{lipsum}

\usepackage[breaklinks=true,bookmarks=false]{hyperref}

\iccvfinalcopy 

\newcommand\blfootnote[1]{%
  \begingroup
  \renewcommand\thefootnote{}\footnotetext{#1}%
  \addtocounter{footnote}{-1}%
  \endgroup
}

\def\iccvPaperID{10767} 

\ificcvfinal\fi

\begin{document}

\title{Efficient-VQGAN: Towards High-Resolution Image \\ Generation  with Efficient Vision Transformers}

\author{
Shiyue Cao$^{1,2*}$,
Yueqin Yin$^{1,2*}$,
Lianghua Huang$^3$,
Yu Liu$^3$,
Xin Zhao$^{1,2,4}$\textsuperscript{\dag}, 
Deli Zhao$^3$,
Kaiqi Huang$^{1,2,4}$ 
\\
$^1$School of Artificial Intelligence, University of Chinese Academy of Sciences
\\
$^2$Institute of Automation, Chinese Academy of Sciences, China
\\
$^3$Machine Intelligence Technology Lab,Alibaba Group, China
\\
$^4$CAS Center for Excellence in Brain Science and Intelligence Technology, China
\\
{\tt\small caoshiyue2021@ia.ac.cn, \{xuangen.hlh, ly103369\}alibaba-inc.com,} \\
{\tt\small xzhaopersonal@foxmail.com,  \{yinyueqin0314, zhaodeli\}@gmail.com, kqhuang@nlpr.ia.ac.cn 
}
}


\maketitle
\ificcvfinal\thispagestyle{empty}\fi

\begin{abstract}
   Vector-quantized image modeling has shown great potential in synthesizing high-quality images. However, generating high-resolution images remains a challenging task due to the quadratic computational overhead of the self-attention process. In this study, we seek to explore a more efficient two-stage framework for high-resolution image generation with improvements in the following three aspects. (1) Based on the observation that the first quantization stage has solid local property, we employ a local attention-based quantization model instead of the global attention mechanism used in previous methods, leading to better efficiency and reconstruction quality. (2) We emphasize the importance of multi-grained feature interaction during image generation and introduce an efficient attention mechanism that combines global attention (long-range semantic consistency within the whole image) and local attention (fined-grained details). This approach results in faster generation speed, higher generation fidelity, and improved resolution. (3) We propose a new generation pipeline incorporating autoencoding training and autoregressive generation strategy, demonstrating a better paradigm for image synthesis.
   Extensive experiments demonstrate the superiority of our approach in high-quality and high-resolution image reconstruction and generation.
   \blfootnote{\hspace{-0.2cm}$^*$  This work was done during an internship at Machine Intelligence Technology Lab, Alibaba Group.}
   \blfootnote{\hspace{-0.2cm}\textsuperscript{\dag}  Corresponding author.}
\end{abstract}

\section{Introduction}

High-fidelity image synthesis has achieved promising performance thanks to the progress of generative models, such as generative adversarial networks (GANs)~\cite{goodfellow2014generative,karras2019style,karras2020analyzing}, diffusion models~\cite{ho2020denoising,dhariwal2021diffusion} and autoregressive models~\cite{esser2021taming,yu2021vector}.
Moreover, high-resolution image generation, a vital generation task with many practical applications, provides better visual effects and user experience in the advertising and design industries. 
Some recent studies have attempted to achieve high-resolution image generation. StyleGAN~\cite{karras2019style,karras2020analyzing} leverages progressive growth to generate high-resolution images. However, GAN-based models often suffer from training stability and poor mode coverage~\cite{salimans2016improved, zhao2018bias}.
As diffusion models continue to evolve, recent studies~\cite{ramesh2022hierarchical, saharia2022photorealistic} have begun to explore the utilization of cascaded diffusion models for generating high-resolution images. This approach involves training multiple independent and  enormous models to collectively accomplish a generation task.
On another note, some researchers~\cite{esser2021taming,yu2021vector,chang2022maskgit} leverage a two-stage vector-quantized (VQ) framework for image generation, which first quantizes images into discrete latent codes and then model the data distribution over the discrete space in the second stage.
Nonetheless, under the limited computational resources (\eg, memory and training time), the architectures of the existing vector-quantized methods are inferior.
In this paper, to solve the problems of existing models, we would like to explore a more efficient two-stage vector quantized framework for high-resolution image generation and make improvements from the following three aspects.

Firstly, prior methods~\cite{esser2021taming,yu2021vector} claim the importance of the attention mechanism in the first quantization stage for better image understanding, and they leverage global attention to capture long-range interactions between discrete tokens. 
However, we find this global attention not necessary for image quantization based on the observation that the alteration of several tokens will only influence their nearby tokens. Hence, local attention can yield satisfactory reconstruction results and circumvent the computationally intensive nature of global attention, especially when generating high-resolution images. 
Consequently, we propose Efficient-VQGAN for image quantization adopting image feature extractor with local attention mechanism. This contributes to the acceleration of image reconstruction and dedicates more computation to the  local information, further improving the reconstruction quality.

Besides, for the second stage of the existing vector-quantized methods~\cite{esser2021taming,yu2021vector,chang2022maskgit}, it would be intractable to generate high-resolution images since the quadratic space and time complexity is respected to the discrete sequence length. Further, the global self-attention interaction could lead to the insufficient ability to capture fine details in local areas. Accordingly, the fined-grained local attention at a token level for better local details capturing plays an essential role as coarse-grained global interaction for long-range context information capturing. We then utilize multi-grained attention, which implements different granularity of attention operations depending on the distance between tokens. As a result, it can support high-resolution image generation with a reduced length of the quantized image token sequence and reasonable computational cost.

Additionally, some recent studies related to text generation~\cite{yang2019xlnet,bi2020palm} in the field of natural language processing, which combine the merits of autoencoding pretraining and autoregressive generation, show great potential in generating high-quality text sequence.
Pretrained autoencoding models like BERT~\cite{kenton2019bert} can exploit bidirectional context to capture more information for reconstructing the masked input corpus, while autoregressive generation performing explicit density estimation can ensure consistency of output token sequence. 
Inspired by such combined training and inference strategy, we propose a similar pipeline for image generation tasks. In the training stage, we utilize an autoencoding-based masked visual token modeling strategy which is trained to recover the randomly masked image tokens by attending to tokens from all directions, better capturing contextual information. In the inference stage, combined with our block-based multi-grained attention mechanism, we autoregressively sample each image block in a fixed order and iteratively sample the tokens within the block in parallel, contributing to improved sampling speed and generation quality. 

The contributions of this work can be summarized as follows.
(1) We propose a more efficient two-stage vector-quantized framework with several improvements in the first quantization stage and the second generative modeling stage, yielding faster computational efficiency and better image quality. (2) We propose a new image generation pipeline that combines the advantages of autoencoding training and autoregressive generation, further improving the synthesis quality.
(3) The proposed two-stage vector-quantized model demonstrates the capability to generate higher-quality images at a faster speed on FFHQ and ImageNet datasets compared to previous methods.

\section{Related Work} \label{Related Work}

\subsection{Image Synthesis}

Recent development in generative modeling enables algorithms to generate high-quality and realistic images.
Generative adversarial networks (GANs) facilitate image generation with promising results. 
However, GAN-based models~\cite{goodfellow2014generative,karras2019style,karras2020analyzing} have poor mode coverage and struggle to model complex distributions due to the inductive priors imposed by convolutions.
Diffusion models have received substantial attention these days, among which many attempts have been made at continuous diffusion models with remarkable results in image generation~\cite{dhariwal2021diffusion,nichol2021improved} and image editing tasks~\cite{kim2022diffusionclip,avrahami2022blended}.
Some Transformer-based methods~\cite{chen2020generative,van2017neural,parmar2018image,razavi2019generating} have shown strong power of density estimation using a fixed forward factorization order for image generation tasks.  GAN-based methods\cite{zhao2021improved,jiang2021transgan} incorporate Transformer with block-wise self-attention to scale up to higher-resolution images. In this paper, we are interested in exploiting two-stage vector quantized models based on vision transformers with multi-grained attention mechanism, which model the data distribution in compressed space and can increase the resolution of the generated images.

\subsection{Vector-Quantized Image Modeling}
Two-stage vector-quantized approaches will first utilize an image tokenizer to extract a discrete image token sequence. Then in the second stage, a generative model is trained to model the token distribution in the discrete latent space.
VQ-VAE~\cite{van2017neural} tokenizes an image into discrete visual tokens by performing online clustering and then models token distribution autoregressively with a convolutional architecture. 
DALL$\cdot$E~\cite{ramesh2021zero} utilizes the first stage of VQ-VAE with Gumbel-Softmax strategy~\cite{jang2016categorical}, and then uses an autoregressive Transformer-based likelihood model to generate visual tokens from the given text input.
VQGAN~\cite{esser2021taming} extends VQ-VAE by adding an adversarial and perceptual metric training loss in the first stage, producing higher-quality reconstructed images.
Recently, ViT-VQGAN~\cite{yu2021vector} improves VQGAN in architecture design and proposes to use a VIT backbone, yielding better reconstruction quality and inefficient. 
However, all those previous approaches employ an autoregressive model for image generation following the raster scan order, which is not efficient.
MaskGIT~\cite{chang2022maskgit} proposes a new image generation paradigm using a bidirectional Transformer decoder trained on Masked Visual Token Modeling(MVTM).
In the inference stage, MaskGIT employs a non-autoregressive decoding strategy to produce images in a constant number of steps.
However, these vector-quantized models still struggle to generate high-resolution images due to the design of sub-optimal two-stage architecture.
Our model develops a more efficient framework for both stages, supporting the generation of high-resolution images.

\begin{figure}[t]
  \centering
   \includegraphics[width=1.0\linewidth]{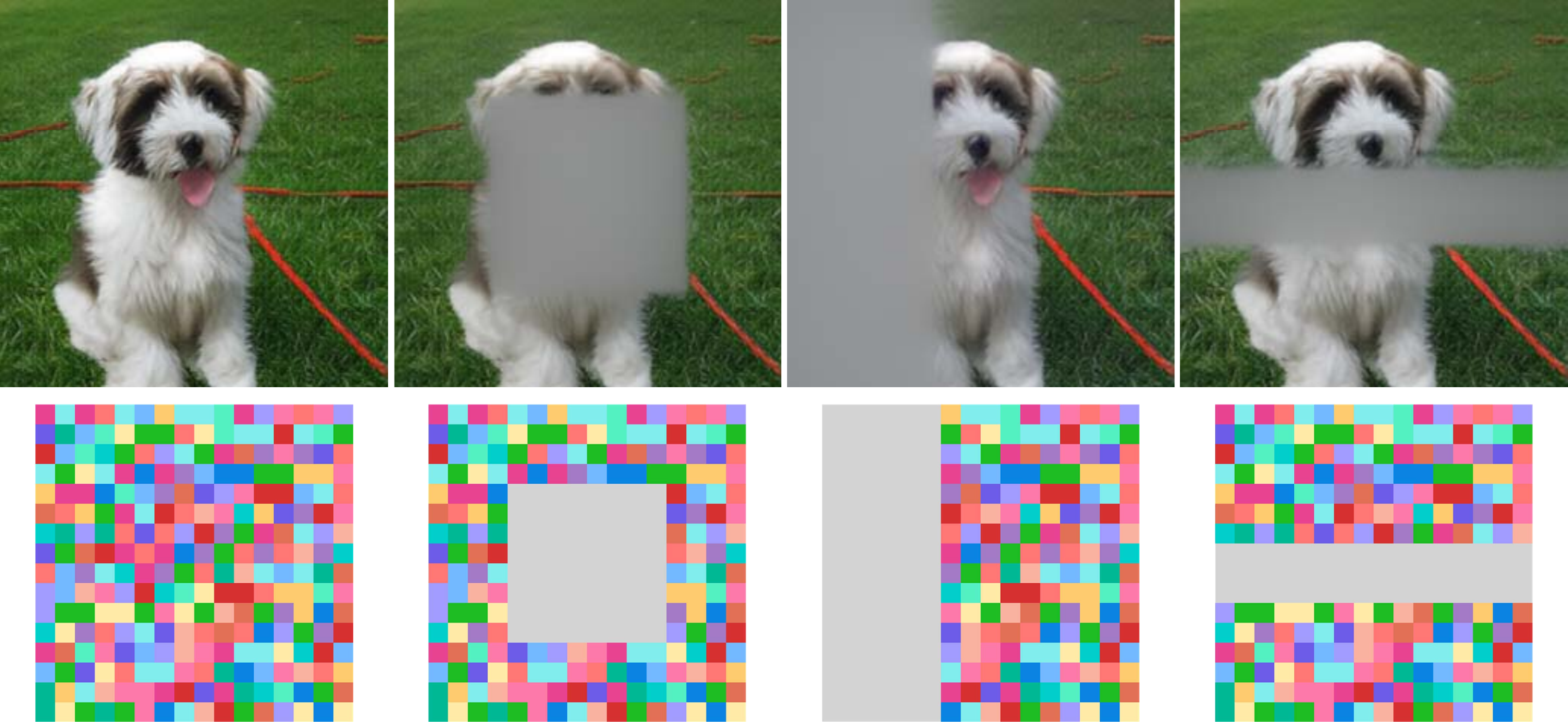}
   \caption{\textbf{The locality of image quantization.} When replacing a local region of latent codes encoded by ViT-VQGAN~\cite{yu2021vector} with a specific token, only surrounding area is affected, while others remain the same. 
   }
   \label{fig:recon_local}
      \vspace{-0.3cm}
\end{figure}

\begin{figure*}[h!]
  \centering
    \vspace{-0.15cm}
    \includegraphics[width=\textwidth]{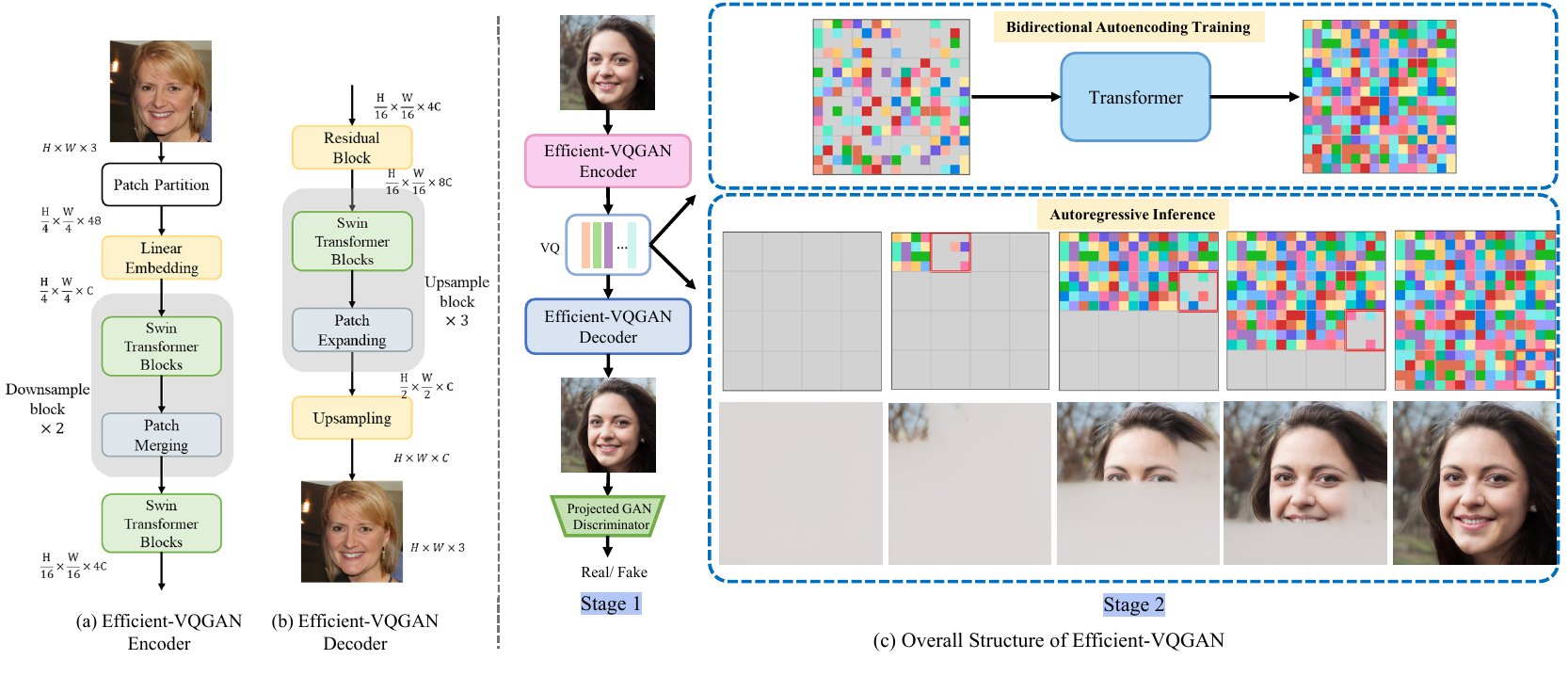}
    \caption{\textbf{ Overview of Efficient-VQGAN.} Efficient-VQGAN consists of two stages: an encoder-decoder-based vector quantization model (left) and a proposed efficient Transformer model. (c) shows the model architecture and pipeline of Efficient-VQGAN. VQ denotes Vector Quantizer. During the training stage, a subset of tokens is randomly replaced by a mask token (marked in gray) and the Transformer is trained to reconstruct them in a masked autoencoding manner. During the generation stage, starting with masked codes, the model gradually predicts tokens block by block. Within each block, the tokens are iteratively sampled parallelly in a few steps.}
   \label{fig:model_architecture}
      \vspace{-0.3cm}

\end{figure*}

\section{Methodology}

Our goal is to achieve high-resolution image generation through a more efficient two-stage vector-quantized image modeling framework, Efficient-VQGAN. Fig.~\ref{fig:model_architecture} shows the structure of our model.
For the first stage, instead of representing an image based on global attention, we design a more efficient vector-quantized model utilizing local attention-based encoder-decoder, as described in Sec.~\ref{locality of image quantization model}.
For the second stage, we propose to learn a Transformer that incorporates global and local attention, which dramatically reduces the sequence length, tackling the quadratic dilemma in previous Transformer-based generative models, as described in Sec.~\ref{Multi-Grained Attention}.
Additionally, we introduce a new training and inference paradigm for image synthesis, which performs masked visual token modeling in the training stage and autoregressive sampling in the generation stage, further improving the image quality, as described in Sec.~\ref{Training and Inference Strategy}.

\subsection{Locality of Image Quantization} \label{locality of image quantization model}

VQGAN~\cite{esser2021taming} adds a non-local attention block in the encoder and decoder model, demonstrating the importance of the attention mechanism for better image understanding.
ViT-VQGAN~\cite{yu2021vector} replace the CNN encoder and decoder with Vision Transformer~\cite{dosovitskiy2020image} using the global attention mechanism, which further improves the reconstruction quality.
However, after conducting extensive experiments, we found that the global attention used in the previous image quantization process is not necessary because of the locality of image quantization (see Fig.~\ref{fig:recon_local}), which instead increases the computational cost.
Based on the observation, we propose our efficient vector-quantized autoencoder, performing local interaction based on Swin Transformer block~\cite{liu2021swin} that has also shown outstanding performance in other image recognition and dense prediction  tasks~\cite{liang2021swinir,cao2021swin}.
The overall architecture of the first stage consists of an encoder, a discrete codebook, a decoder, and a discriminator. 
Given an image $\bm{x} \in \mathbb{R}^{3\times H \times W}$, the encoder will downsample it into a feature map $\bm{z} \in \mathbb{R}^{d \times \frac{H}{16} \times \frac{W}{16}}$. 
Then a discrete codebook is queried to produce a quantized image feature map $z_q$, and then we feed $z_q$ to the decoder to reconstruct the original image. 


\noindent \textbf{Efficient-VQGAN Encoder.} 
Given an input image, the encoder will first divide it into patches with fixed size ($4 \times 4$), and then a linear embedding layer is applied to transform the patch feature into an arbitrary dimension.
Then the patch feature will be passed through several Swin Transformer blocks to perform local attention, and achieve $2\times$ downsampling through patch merging layers. Downsampling factor $f$ can be adjusted by adding or removing downsample blocks.
Therefore, after being encoded by the Efficient-VQGAN encoder with default two downsample blocks, the resolution of the feature map is $1/16$ of the original image.  Output feature map $z$ produced by encoder will be passed through the quantized module introduced from VQGAN~\cite{esser2021taming} to output quantized feature map $z_q$.

\noindent \textbf{Efficient-VQGAN Decoder.} The Efficient-VQGAN decoder and encoder are symmetrical.
The decoder module also consists of several stages, totally implement upsample scale equal to $f$.
In these stages, each Patch Expanding block achieve $2 \times$ upsampling, 
and finally, a nearest neighbor interpolation upsample module is used to perform $2 \times$ up-sampling to generate the reconstructed image. 

\noindent \textbf{Training loss of Image Quantization.} 
To enable training process stable and convergent for  high-resolution images, Projected GANs discriminator~\cite{sauer2021projected} is applied to produce GAN loss $L_{Adv}$, discriminating samples in deep feature space by using pretrained image feature extraction model, leading to improvement in quality and convergence speed. Besides, perceptual loss~\cite{johnson2016perceptual} is applied to enhance details and perceptual quality.
The pixel level $l_2$ loss between input and reconstructed images and vector quantization loss~\cite{esser2021taming} $L_{VQ}$ are also introduced. The total loss defined as follow:
\begin{equation}
L=L_{Perceptual}+L_2+L_{VQ}+\lambda*L_{Adv} \\
\end{equation}
where  $ \lambda $ is adaptive weight~\cite{esser2021taming}.


\begin{figure}[t]
  \centering
   \includegraphics[width=\linewidth]{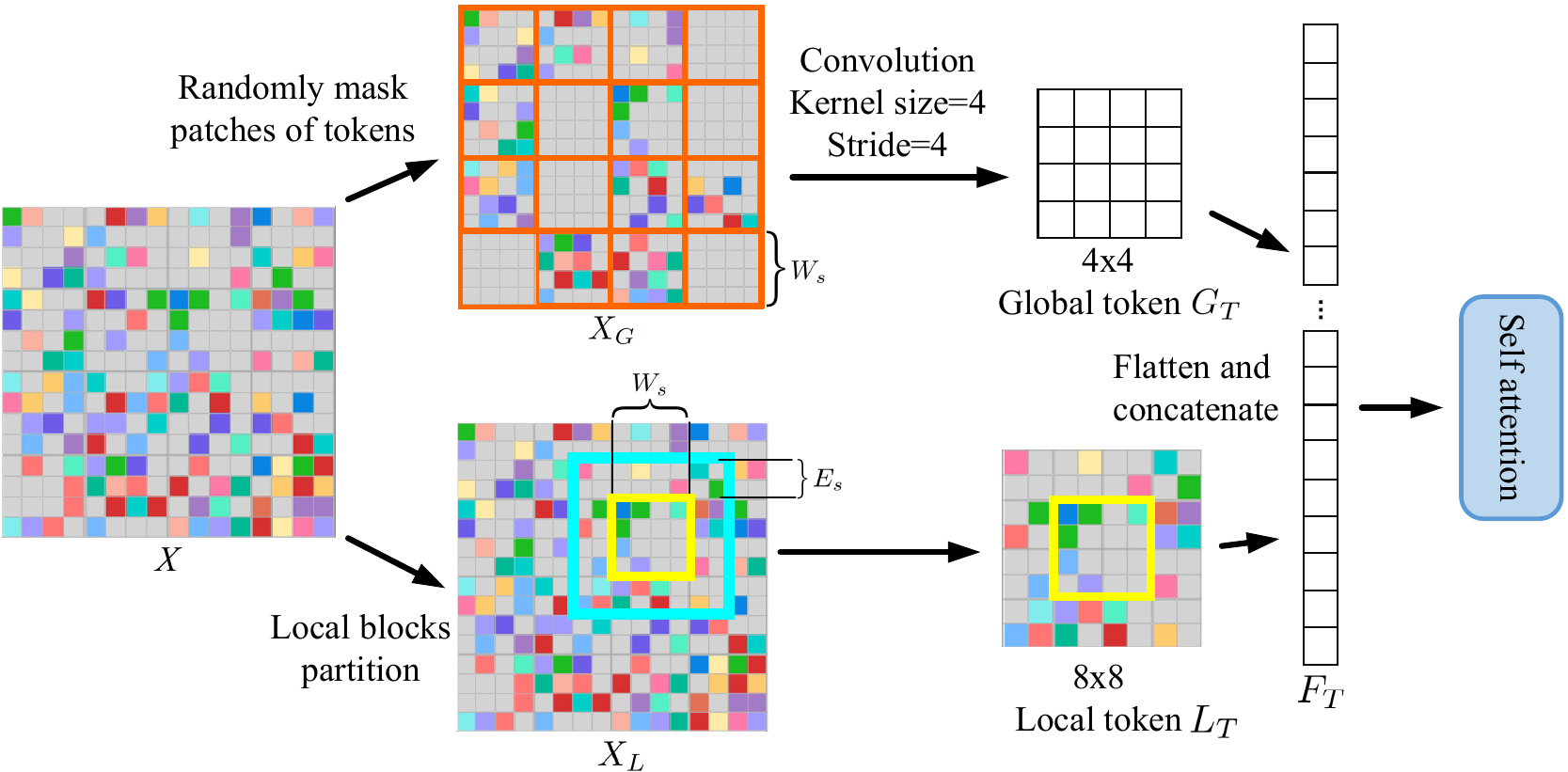}
   \caption{\textbf{Multi-Grained Attention calculating process.} Yellow square in $X_L$ denotes the modeled target tokens. Blue square in $X_L$ denotes the extended local token matrix. Each local block will be transformed into a global token (red square in $X_G$).
   }
   \label{fig:_multi-grained}
   \vspace{-0.3cm}
\end{figure}

\subsection{Multi-Grained Attention for Efficient Image Generation} \label{Multi-Grained Attention}

The quadratic computational complexity of the number of visual tokens limits the capability of self-attention Transformer based-models to generate high-resolution images. As mentioned in Sec.~\ref{locality of image quantization model}, applying global attention to each visual token is redundant and costly. Nevertheless, we cannot naively employ only local attention again to reduce  complexity, since the generation process differs from image quantization, where global attention maintains the semantic consistency of the image. Consequently, both global and local attention is required to model the visual tokens by the transformer. 

Inspired by some works~\cite{zhang2021multi,yang2021focal} which leverages the multi-grained attention mechanism in image recognition task, we combine this mechanism with our masking strategy for image generation. Given an image $I$, we can obtain the discrete token matrix $X = (x_1, x_2, \cdots, x_{H \times W})$ by passing the image through the Efficient-VQGAN encoder and quantization module, where $H,W$ is the latent size.
Then the image token matrix is split into blocks of size $W_s \times W_s$, as shown in Fig.~\ref{fig:_multi-grained}. 
We then perform the multi-grained attention for each query token in the token matrix.
Here, we define the query token (the token within yellow square in $X_L$) as the target tokens will be predicted in this calculation. Then we define the extend local tokens (blue square in $X_L$ ) which is the region surrounding the query block with a block size of $ (W_s + 2E_s) \times (W_s + 2E_s)$ to provide more local information cross the blocks, where $E_s$ denotes extend size. As for the global tokens $G_T$, we separately perform a convolutional operation for each block to group all the tokens in a block into a global token.
Here, the filter size and convolutional stride are set to the block size $W_s$.
After this aggregation operation, we obtain $\frac{H}{W_s} \times \frac{W}{W_s}$ global tokens for a token matrix.
The final token sequence is defined as the concatenation of global tokens and local tokens, namely $F_T =[G_T, L_T]$.
We then feed $F_T$ into Transformer to perform self-attention and output the refined feature that incorporates both the global semantic and local contextual information.

The design of combining block-based global interaction, with local interaction calculated within a local block, leading to considerable reduction in the number of tokens, which facilitate our model to achieve efficient high-resolution image generation.

\subsection{Autoencoding Training and Autoregressive Inference} \label{Training and Inference Strategy}

In natural language generation tasks, some researchers~\cite{yang2019xlnet,bi2020palm} point out the pros and cons of BERT-like autoencoding training and autoregressive language modeling.
These works aim at exploring a pretraining objective that combines the advantages of both strategies while avoiding their weaknesses.
In general, both natural language generation and VQ-based image generation aim at finding optimal token sequence $s = (s_1, s_2, \cdots, s_l)$. When adopting the autoregressive generation strategy, we can factorize the probability likelihood into a forward product, which can be learned readily by the model $p_{\theta}(s) = \prod_{i=1}^{l} p_{\theta}^i(s_i|s_{<i})$.
Nonetheless, autoregressive models are challenging to capture deep bidirectional context information due to the uni-directional training constraint. On the other hand, BERT-like bidirectional autoencoding training strategy allows the model to capture bidirectional context better, however, an independent assumption is required that the tokens predicted in parallel ought to be independent of each other, otherwise semantic inconsistency will occur.
Accordingly, in this section, we propose our training and inference pipeline, which incorporates the unique advantages of autoencoding training and autoregressive generation.

\noindent \textbf{Masked Autoencoding training Strategy.} Given the discrete token sequence $X = (x_1, x_2, \cdots, x_{H \times W})$, following MaskGIT~\cite{chang2022maskgit}, a subset of tokens are randomly masked.
The mask ratio is defined by a cosine scheduling function $\gamma (r) = \text{cosine}(\frac{\pi}{2}r) \in (0,1]$, where the ratio $r$ is from $0$ to $1$. 
 We uniformly choose $\lceil \gamma(r) \cdot (H \times W) \rceil$ tokens in $X$ and replace them with mask token, producing the corrupted token matrix $X_{L} = (x_1, [\text{\tt{MASK}}] , \cdots, [\text{\tt{MASK}}], x_{H \times W})$.

During the training process, we divide the masking strategy into two parts, for calculating local and global attention, respectively.
Local attention is performed on $X_{L}$ within each block as mentioned above to produce $L_T$.
When calculating global attention, entire patches of tokens will be randomly masked first. This operation enables our model to predict blocks in arbitrary order, including autoregressive order and non-autoregressive order, which also make it possible to achieve inpainting, outpainting and image editing tasks.
Then we perform a convolutional operation that transforms each block in $X_{G}$ into one global token, producing a corrupted global token matrix with $\frac{H}{W_s} \times \frac{W}{W_s}$ global visual tokens, namely $ G_{T}$.
After obtaining all tokens $F_T$, we feed them to the Transformer module to predict the probability distribution of each masked token within one block.
The loss function can be formulated as:
\begin{equation}
  \mathcal{L} = - \mathop{\mathbb{E}} \Big[ \sum_{\forall i \in [1,H \times W], L_{T_i}= [\text{\tt{MASK}}]} \log p(x_i| L_T, G_T) \Big],
\end{equation}
where the negative log-likelihood is computed as the cross-entropy reconstruction loss between the true one-hot token within a block and the predicted token.

\noindent \textbf{Autoregressive Inference Strategy.} 
In previous autoregressive decoding methods~\cite{esser2021taming,yu2021vector}, tokens are sequentially generated based on all previously generated tokens, which could improve the generation consistency. For high-resolution image generation, however, the sampling speed is intolerable due to the long sequence length. 
Based on bidirectional training, MaskGIT~\cite{chang2022maskgit} can generate multiple image tokens in a single pass, and iteratively generate a complete image, which greatly reduces the sampling steps during inference.
However, as claimed in ~\cite{tang2022improved}, when sampling multiple tokens simultaneously and each token is sampled independently with an estimated probability will result in ignoring the dependencies between different tokens at different locations.
In ~\cite{tang2022improved}, they propose a fewer tokens sampling strategy that samples fewer tokens at each step to alleviate this joint distribution issue.
Inspired by such strategy, we propose to generate image tokens block by block in an autoregressive manner, which incorporates the design of our block-based model architecture and the merit of autoregressive sampling.
The difference compared to ~\cite{tang2022improved} is that ~\cite{tang2022improved} fixes the number of tokens generated in the whole image for each step, we fix the order of generation for each step, which essentially achieves a similar effect.
Besides, in each block, we adopt parallel decoding used in MaskGIT that generates multiple tokens at the same time within a block.
Under the setting of MaskGIT training strategy that randomly masks tokens in a whole image, it is difficult to include all the situations during inference stage (different masked token mode), especially for high-resolution image generation.
However, limiting parallel sampling to a single block in our inference strategy would greatly alleviate this problem.
Therefore, our sampling strategy follows an overall autoregressive and local parallel generation manner.

\section{Experiments}

\begin{figure}[t]
    \centering

   \includegraphics[width=\linewidth]{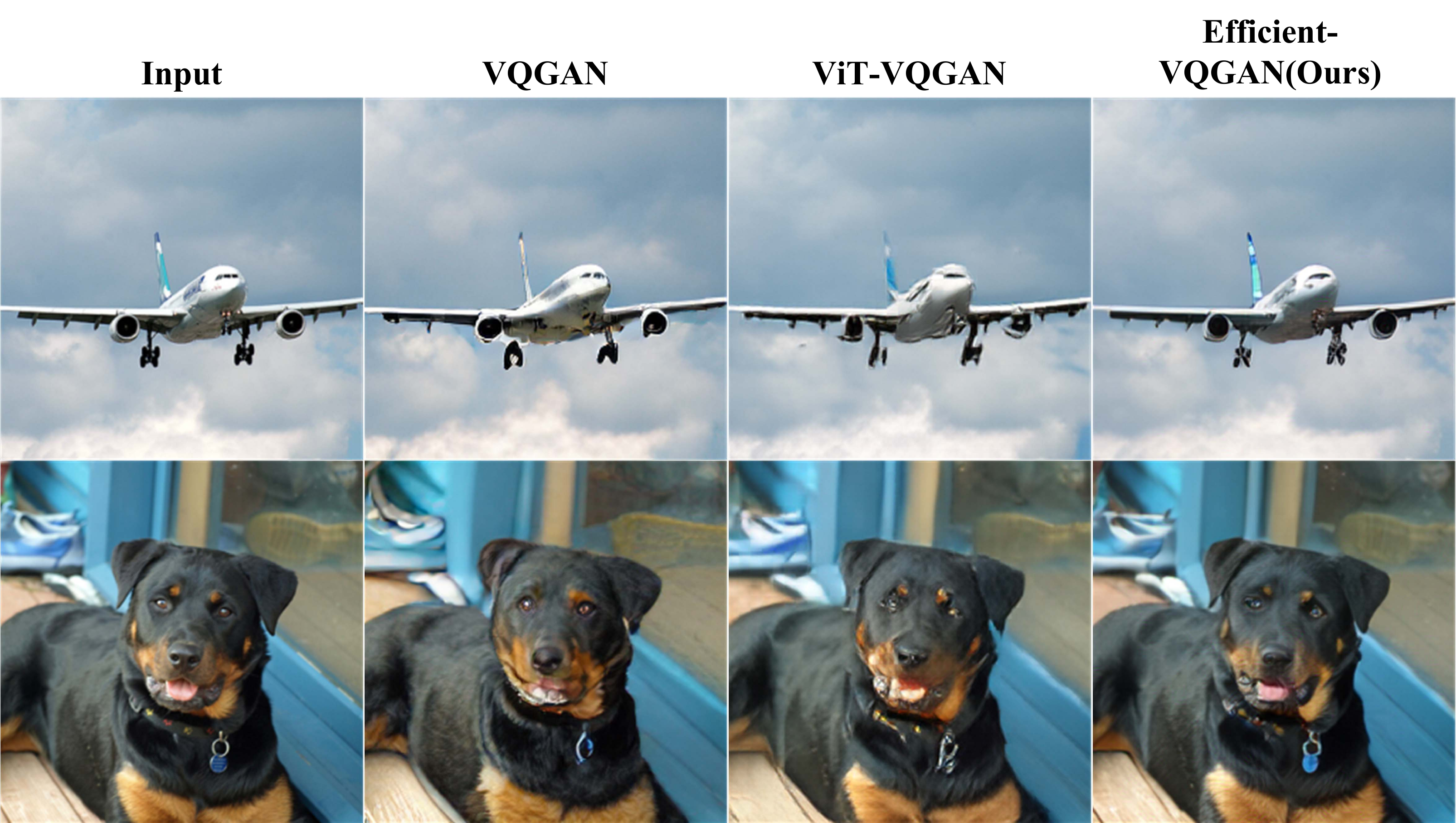}
   \caption{\textbf{Reconstruction comparison between VQGAN~\cite{esser2021taming}, ViT-VQGAN~\cite{yu2021vector} and Efficient-VQGAN on ImageNet dataset.} Ours can perfectly reconstruct the original image, preserving more details compared to others.
   }
   \label{fig:vqgan_recon_compare}
   \vspace{-0.3cm}
\end{figure}

In this section, we evaluate the ability of Efficient-VQGAN in the first vector quantization stage (Sec.~\ref{Quantization}) and image synthesis tasks (Sec.~\ref{Synthesis}) on FFHQ~\cite{karras2019style}, CelebA-HQ~\cite{karras2017progressive}, and ImageNet~\cite{deng2009imagenet} dataset. In Sec.~\ref{Editing}, we present some direct applications of Efficient-VQGAN on image inpainting, outpainting, and editing tasks. In Sec.~\ref{Ablation}, we conduct some ablation studies. The following tests are implemented on $256^2$ images unless otherwise specified. Details of experiment settings can be found in our appendix.

\begin{table}[h!]
  \begin{center}
  \tabcolsep=0.1cm
  \scalebox{0.80}{
  \begin{tabular}{lcccc}
  \hline
  Methods      & Dataset  & \begin{tabular}[c]{@{}c@{}} Codebook \\ Size \end{tabular} & \begin{tabular}[c]{@{}c@{}} Latent \\ Size \end{tabular}   & \begin{tabular}[c]{@{}c@{}}FID on\\ Validation \end{tabular} $\downarrow$ \\ \hline
  VQGAN~\cite{esser2021taming}      & FFHQ     & 1024  & 16x16          & 4.39              \\
  ViT-VQGAN~\cite{yu2021vector}  & FFHQ     & 8192  & 32x32          & 3.13              \\
  Ours & FFHQ     & 1024  & 16x16          & 3.38              \\
  Ours & FFHQ     & 1024  & 32x32          & \textbf{2.72}              \\ \hline
  VQGAN      & ImageNet & 16384 & 16x16          & 4.98              \\
  VQGAN*     & ImageNet & 8192  & 32x32          & 1.49              \\
  VQGAN**    & ImageNet & 512   & 64x64 \& 32x32 & 1.45              \\
  ViT-VQGAN  & ImageNet & 8192  & 32x32          & 1.55              \\
  Ours & ImageNet & 1024  & 16x16          & 2.34              \\
  Ours & ImageNet & 1024  & 32x32          & \textbf{0.95}            \\ \hline
  \end{tabular} }
  \end{center}
  \caption{\textbf{Fréchet Inception Distance (FID)~\cite{heusel2017gans}  between reconstructed validation split and original validation split.} $*$ means model trained with Gumbel-Softmax strategy. $**$ means model leveraging multi-scale hierarchical codebook proposed in ~\cite{jang2016categorical}. Ours shows the best reconstruction quality.}
    \label{tab:recon_fid}
  \end{table}

  \begin{table}[]
    \begin{center}
    \tabcolsep=0.1cm
    \scalebox{0.8}{
    \begin{tabular}{lcccc}
\hline
Methods      & \begin{tabular}[c]{@{}c@{}} Downsampling \\ factor $f$ \end{tabular}   & Latent Size &  \begin{tabular}[c]{@{}c@{}}Throughput\\ (imgs/sec)\end{tabular} $\uparrow$ & FID$\downarrow$  \\ \hline
VQGAN~\cite{esser2021taming}      & 16 & 16x16                                                           & 112                                                             & 4.98 \\
Ours & 16 & 16x16                                                            & \textbf{142}                                                             & \textbf{2.34} \\ \hline
VQGAN~\cite{esser2021taming}      & 8  & 32x32                                                        & 103                                                             & 1.49 \\
ViT-VQGAN~\cite{yu2021vector}  & 8  & 32x32                                                      & 95                                                              & 1.55 \\
Ours & 8  & 32x32                                                           & \textbf{108}                                                             & \textbf{0.95} \\ \hline
\end{tabular} 
}
\end{center}
    \caption{\textbf{Reconstruction speed comparison}. Under the same latent size, ours achieve faster reconstruction speed and better quality. Tests are implemented on single A100-80GB GPU.}
    \label{tab:recon_speed}
    \vspace{-0.3cm}
    \end{table}

\begin{figure*}[t]
\centering
  \includegraphics[width=0.95 \linewidth]{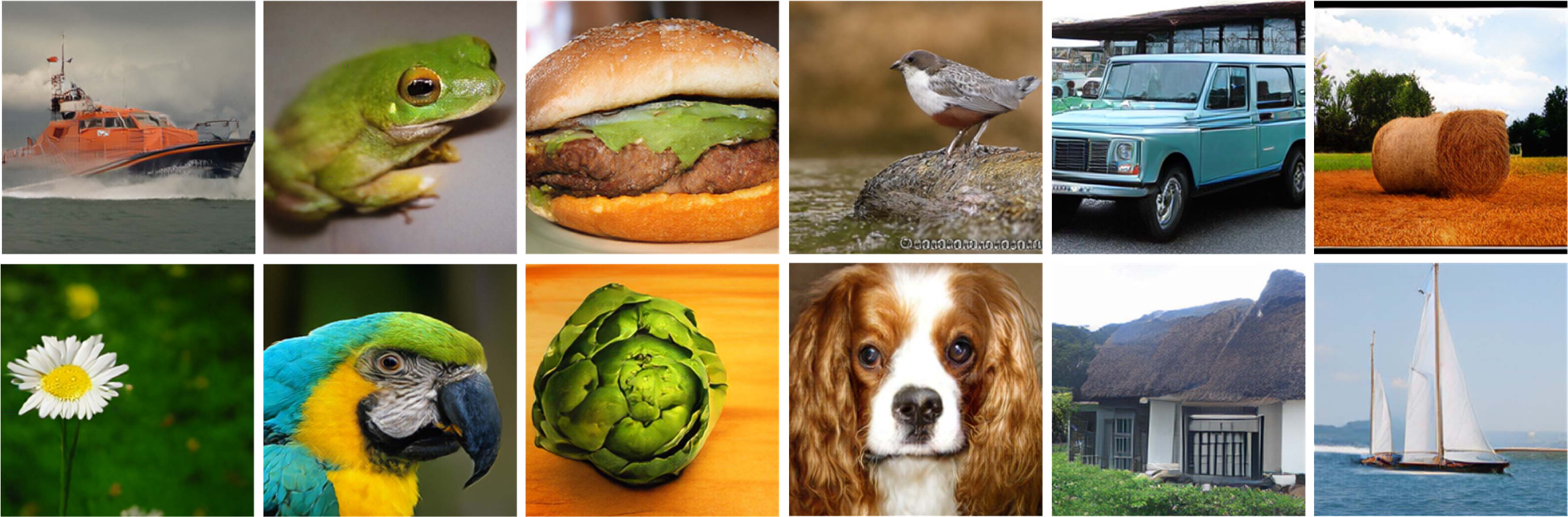}

  \caption{\textbf{Synthesized samples by Efficient-VQGAN on ImageNet dataset at $256^2$ resolution.}
  }
  \label{fig:imagenet_sample}
  \vspace{-0.3cm}
\end{figure*}

\subsection{Image Quantization}
\label{Quantization}


When training the image quantization model, we follow the default train and validation split for each dataset and conduct experiments with different downsampling factor $f$.
For each setting, the Codebook Size $|Z|$ is set to $1024$. 
On FFHQ and CelebA-HQ dataset, our models are trained with a batch size of $16$ in 8*A100 GPU for a total of $50$ epochs, while on ImageNet we find that the model has converged for only $20$ epochs.
Quantitative reconstruction comparison results ($256^2$ resolution) are shown in Tab.~\ref{tab:recon_fid} and Tab.~\ref{tab:recon_speed}. 
Our method achieves higher inference speed due to the reduced computation cost of local attention operations. Besides, under the same downsampling factor $f$, our model achieves better reconstruction fidelity with a lower capacity of codebook ($1024$ codebook entries). Qualitative comparison results display in Fig.~\ref{fig:vqgan_recon_compare}.
We claim that the global attention mechanism may bring some noises during image quantization so that our method which only performs local attention could put more computation on local interactions and improve generative details. More quantitative results for different resolutions ($512^2, 1024^2$) can be found in our appendix. 




\begin{figure}[t]
    \centering

   \includegraphics[width=1.0\linewidth]{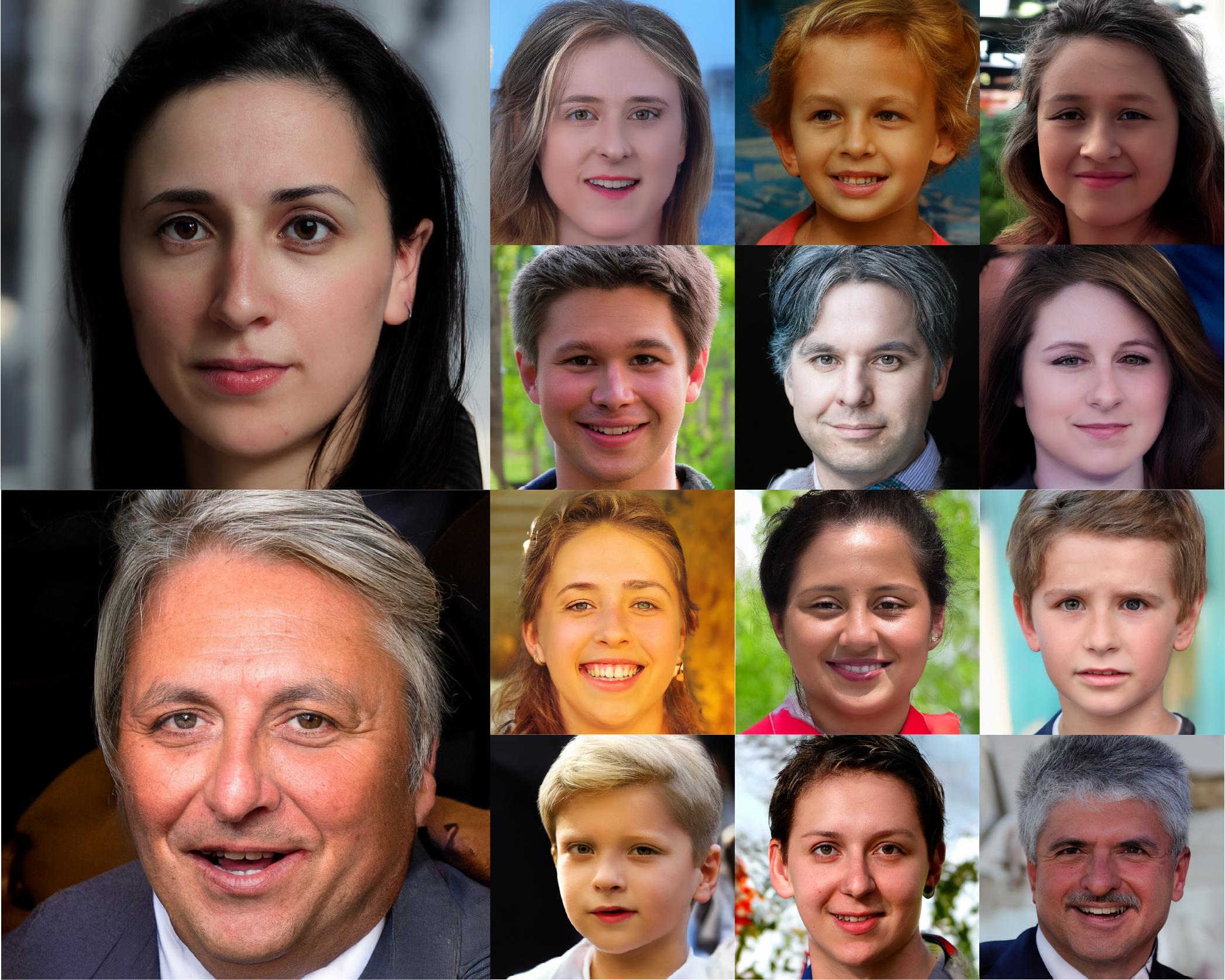}

   \caption{\textbf{Synthesized samples by Efficient-VQGAN on FFHQ dataset at $1024^2$ resolution.} Best viewed zoomed in.
   }
   \label{fig:1024_ffhq}
   \vspace{-0.5cm}
\end{figure}





\subsection{Image Synthesis}
\label{Synthesis}

\begin{table}[]
\begin{center}
\tabcolsep=0.1cm

\begin{tabular}{lccclcc}
\hline
\multicolumn{3}{c}{CelebA-HQ $256\times256$} &  & \multicolumn{3}{c}{FFHQ  $256\times256$} \\ \cline{1-3} \cline{5-7} 
Methods           &        & FID  $\downarrow$      &  & Methods           &     & FID  $\downarrow$     \\ \cline{1-3} \cline{5-7} 
NVAE~\cite{vahdat2020nvae}              &        & 40.3      &  & BigGAN~\cite{brock2018large}           &     & 12.4    \\
VAEBM~\cite{xiao2020vaebm}            &        & 20.4      &  & ImageBART~\cite{esser2021imagebart}        &     & 9.57    \\
Style ALAE~\cite{pidhorskyi2020adversarial}       &        & 19.2      &  & GANformer~\cite{hudson2021generative}        &     & 7.42    \\
DC-VAE~\cite{parmar2021dual}           &        & 15.8      &  & VQ-Diffusion~\cite{gu2022vector}     &     & 6.33    \\
StyleSwin~\cite{zhang2022styleswin}        &        & 3.25      &  & StyleGAN-XL~\cite{sauer2022stylegan}      &     & 2.19    \\ \cline{1-3} \cline{5-7} 
VQGAN~\cite{esser2021taming}            &        & 10.2      &  & VQGAN~\cite{esser2021taming}            &     & 9.6     \\
VIM-Large~\cite{yu2021vector}        &        & 7.0         &  & VIM-Large~\cite{yu2021vector}        &     & 5.3     \\
Ours             &        & 7.81      &  & Ours             &     & 5.28    \\ \hline
\end{tabular} \end{center}

\caption{ \textbf{Quantitative comparison of face image generation.}}
\label{tab:fid_gen}
\vspace{-0.2cm}
\end{table}

\begin{table}[]
  \begin{center}
\begin{tabular}{lc}
\hline
Methods       & \multicolumn{1}{c}{FID$\downarrow$} \\ \hline
StyleGAN-XL~\cite{sauer2022stylegan}    &	2.02 \\
StyleGAN2~\cite{karras2020analyzing}   & 2.84                    \\
HiT-B~\cite{zhao2021improved}       & 6.37                    \\
Ours        & 11.81                   \\
StyleALAE~\cite{pidhorskyi2020adversarial}   & 13.09                   \\ \hline
\end{tabular} \end{center}
\caption{\textbf{Quantitative comparison on FFHQ at \bf{$1024^2$}  resolution.}
}
\label{tab:ffhq_1024}
\vspace{-0.6cm}
\end{table}

Based on the well-trained Efficient-VQGAN autoencoder, we train our transformer model and evaluate the performance on unconditional and class-conditioned image synthesis tasks. To speed up the generation process, we set the downsampling factor $f$ of the first stage to $16$.
Our models are trained in 8*A100 GPU for a total of $200$ epochs. 
 When synthesizing samples, in-block iteration steps $T=8$ is applied. 

We compare the quantitative results of our Efficient-VQGAN with several state-of-the-art methods on FID score. The results of unconditional image synthesis on CelebA-HQ and FFHQ are shown in Tab.~\ref{tab:fid_gen}. While some task-specialized GAN models report better FID scores, our model architecture is more flexible and can support a variety of tasks. The reported FID score of VIM-large~\cite{yu2021vector} on CelebA-HQ is slightly better than ours, due to the number of parameters that are 8 times larger than our model. 
Quantitative comparisons of face synthesis at 1024 resolution are shown in Tab.~\ref{tab:ffhq_1024}. Existing methods are almost GAN-based, and to our knowledge, no VQ-based method reports better performance.
Compared to our baseline VQGAN~\cite{esser2021taming} and VIM-base~\cite{yu2021vector}, we also improve the quality of class-conditioned image synthesis as shown in Tab.~\ref{tab:imagenet_fid}.
The impressive generated images of ImageNet dataset are shown in Fig.~\ref{fig:imagenet_sample}. 
Furthermore, we evaluate the memory cost of existing vector quantized transformer models during training time, as shown in Tab.~\ref{tab:memory}.
When the image token sequence is too long, the out-of-memory(OOM) issue occurs in other methods due to the computational overhead of the global self-attention mechanism.
Benefiting from the multi-grained attention design, our model requires less memory resources and can synthesize higher resolution images ($1024^2$), as shown in Fig.~\ref{fig:1024_ffhq}.
We also adopt the pretraind VQGAN quantizer~\cite{esser2021taming} as the first stage of our model and retrained the generative model for the second stage. On the basis of the same quantization model, the generation results of our model greatly exceed the baseline VQGAN model~\cite{esser2021taming} (see Tab.~\ref{tab:stage2compare}), demonstrating the effectiveness of our multi-grained attention and overall generation pipeline.

\begin{table}[]
\begin{center}
\scalebox{0.95}{
\begin{tabular}{lccc}
\hline
Methods        & Acceptance Rate & FID $\downarrow$   & IS $\uparrow$      \\ \hline
IDDPM~\cite{nichol2021improved}        & 1               & 12.3  & -      \\
StyleGAN-XL~\cite{sauer2022stylegan}  & 1               & 2.3   & 265.12 \\
VQ-Diffusion~\cite{gu2022vector} & 1               & 11.89 & -      \\
ADM-G~\cite{dhariwal2021diffusion}        & 1               & 10.94 & 101    \\ \hline
VIM-Base~\cite{yu2021vector}     & 1               & 11.2  & 97.2   \\
MaskGIT~\cite{chang2022maskgit}       & 1               & 6.18  & 182.1  \\
VQGAN~\cite{esser2021taming}        & 1               & 17.04 & 70.6   \\
VQGAN~\cite{esser2021taming}        & 0.5             & 10.26 & 125.5  \\
Ours         & 1               & 9.92  & 82.2   \\
Ours         & 0.5             & 6.81  & 135.84 \\ \hline
\end{tabular}}
\end{center}
\caption{\textbf{FID comparison of class-conditioned image synthesis on ImageNet at \bf{$256^2$} resolution.} All VQ-based models above take $16\times16$ latent size.  Acceptance rate reports classifier-based rejection sampling using ResNet-101.
}
\label{tab:imagenet_fid}
\vspace{-0.3cm}
\end{table}

\begin{table}[]
\begin{center}
  \tabcolsep=0.1cm
  \scalebox{0.9}{
\begin{tabular}{lcccc}
\hline
                &          & \multicolumn{3}{c}{Training Time Memory Cost {[}M{]}  $\downarrow$}                                                                                                                                     \\ \cline{3-5} 
Methods           & \#Params & \begin{tabular}[c]{@{}c@{}}Latent Size\\ 32x32\end{tabular} & \begin{tabular}[c]{@{}c@{}}Latent Size\\ 64x64\end{tabular} & \begin{tabular}[c]{@{}c@{}}Latent Size\\ 128x128\end{tabular} \\ \hline
VQGAN~\cite{esser2021taming}           & 1.4B     & 21848                                                       & OOM                                                         & OOM                                                           \\
VIM-Base~\cite{yu2021vector}        & 650M     & 11638                                                       & OOM                                                         & OOM                                                           \\
MaskGIT~\cite{chang2022maskgit}         & 227M     & 6780                                                        & 49056                                                       & OOM                                                           \\
Ours & 185M     & 5560                                                        & 14154                                                       & 58096                                                         \\ \hline
\end{tabular} 
}
\end{center}
   \caption{\textbf{Training time memory cost of Stage-2 models in different latent size.} For a fixed image quantization compress rate, a larger latent size indicates higher image resolution. OOM denotes out-of-memory. 
   }
   \label{tab:memory}
   \vspace{-0.3cm}
\end{table}

\begin{table}[t]
  \begin{center}
\begin{tabular}{ccc}
\hline
Dataset  & Ours FID $\downarrow$   & Baseline FID~\cite{esser2021taming} $\downarrow$   \\ \hline
FFHQ     & 7.5   & 9.6          \\
ImageNet & 13.67 & 17.04        \\ \hline
\end{tabular} \end{center}
   \caption{\textbf{FID comparison of second stage model between Ours and VQGAN based on the same first stage quantization model.} Proposed model greatly outperforms VQGAN stage-2 baseline. 
   } 
   \label{tab:stage2compare}
   \vspace{-0.3cm}
\end{table}

\subsection{Image Editing Applications}
\label{Editing}

\begin{figure}[t]
    \centering

   \includegraphics[width=1.0\linewidth]{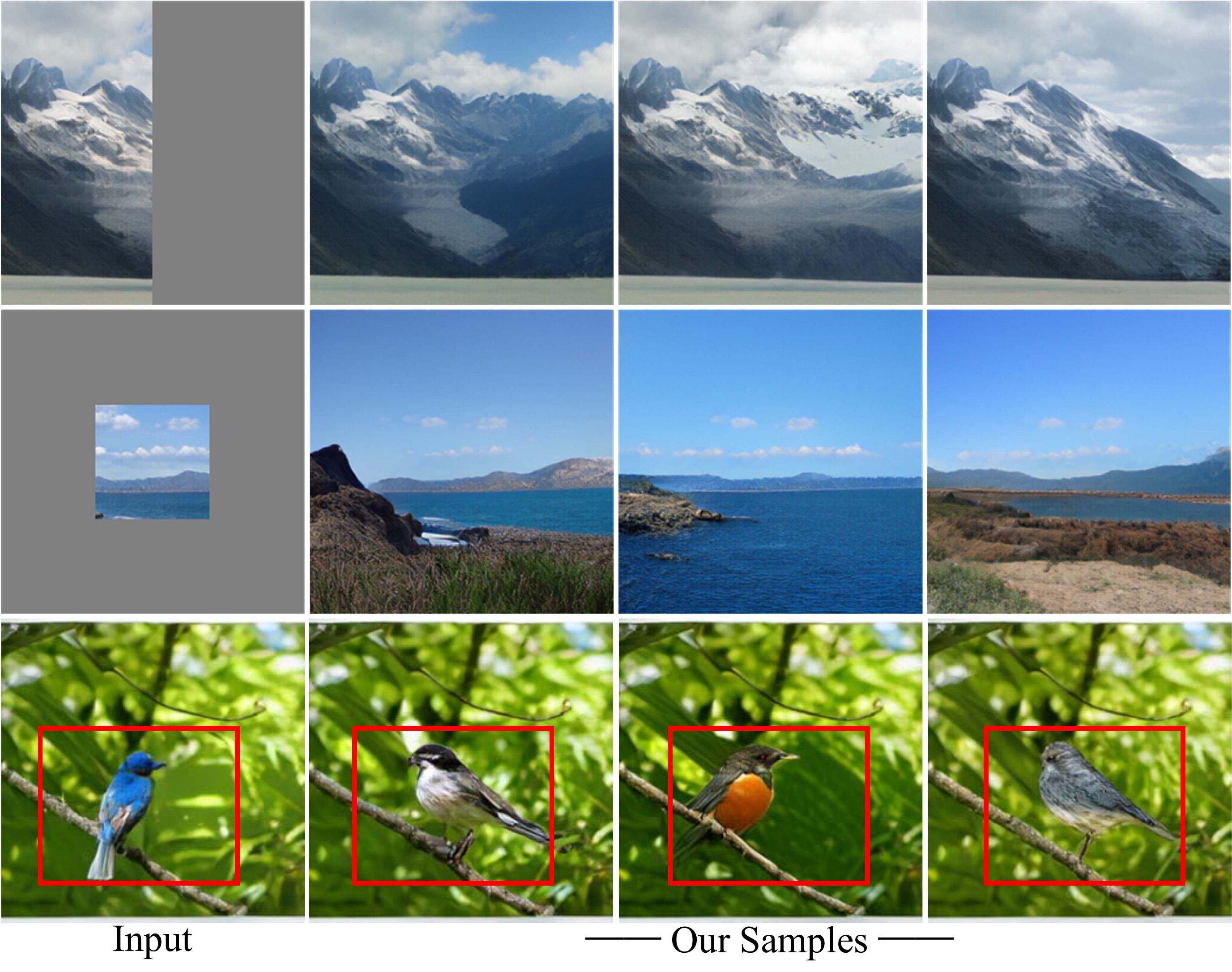}
\vspace{-0.3cm}
   \caption{\textbf{Inpainting, outpainting and class-conditional image editing.} While maintaining the semantic consistency of images, our model shows the diversity of inpainting (first row) and outpainting (second row). Replace the given image in the bounding box with a target class object (third row). 
   }
   \label{fig:inpaint}
   \vspace{-0.6cm}
\end{figure}

Due to causality limitations in the inference process, it is challenging for autoregressive generative models to perform image editing.
Efficient-VQGAN can be seamlessly applied to three image editing tasks (see Fig.~\ref{fig:inpaint}), without any modifications of the model architecture. 
We tokenize the input masked image and feed the corrupted image token matrix into the second stage to iteratively complete the masked image.
As shown in the first and second row of Fig.~\ref{fig:inpaint}, Efficient-VQGAN can make consistent completions thanks to the multi-grained interaction that allows the model to capture both the global semantic as well as the local fine-grained details. Further, due to the randomness in the inference stage, we can obtain diverse editing results.
Class-conditional image editing is defined as regenerating the image content inside a bounding box conditioned on the given class label. Efficient-VQGAN can replace the selected object while preserving the background outside the bounding box, and the entire image is visually harmonious.

\subsection{Ablation Studies}
\label{Ablation}

We conduct ablation studies at $256^2$ resolution images with $16\times16$ latent size. More ablation results could be found in our appendix.

\begin{figure}[t]
    \centering

  \tabcolsep=0.1cm
   \includegraphics[width=0.48\linewidth]{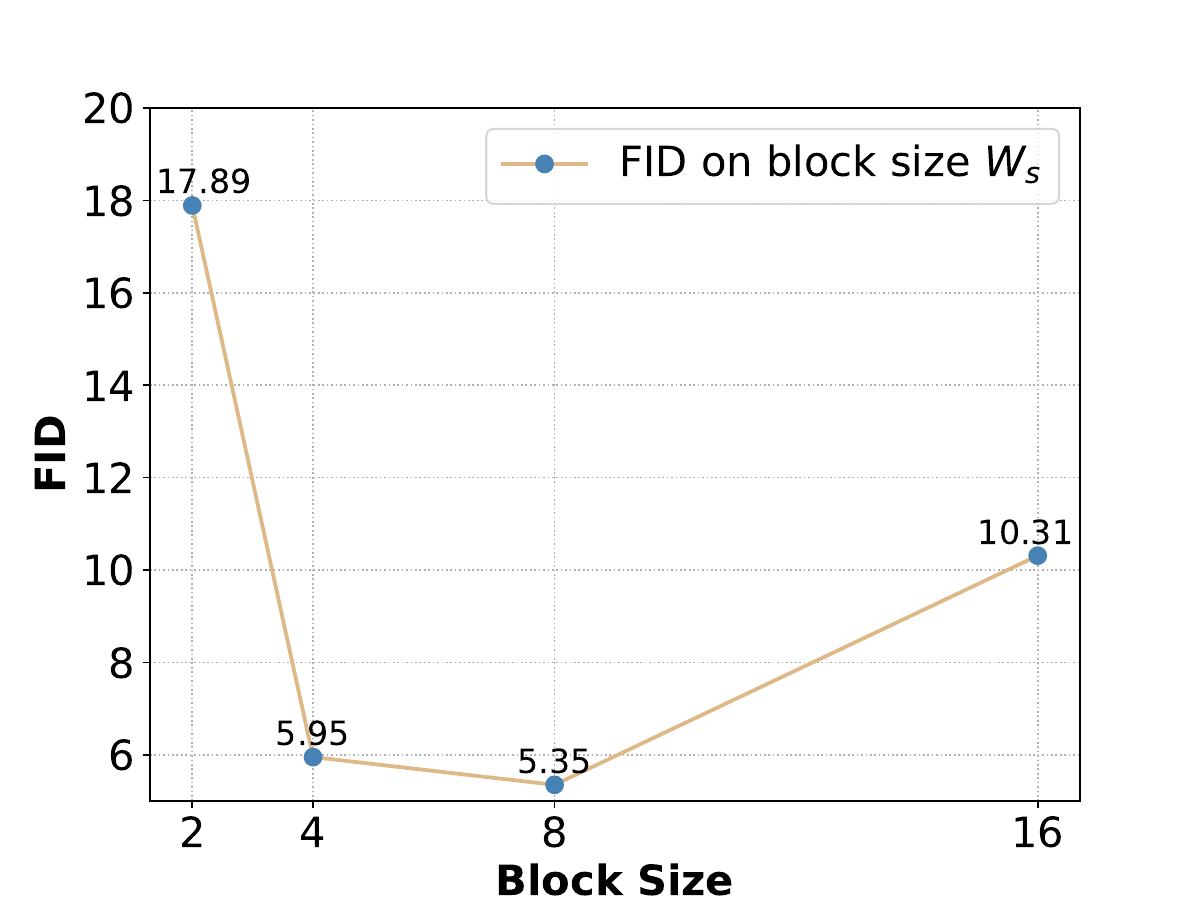}
   \includegraphics[width=0.48\linewidth]{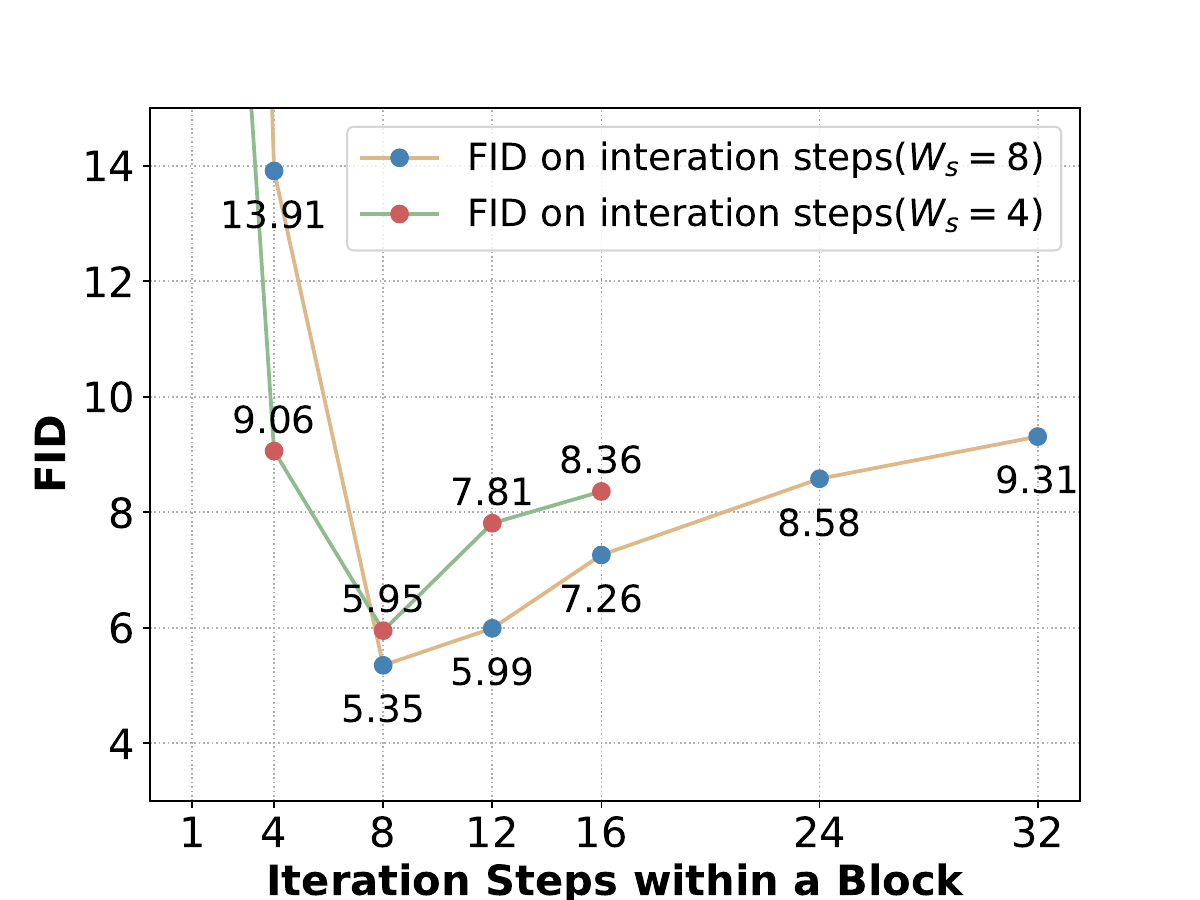}
   \caption{\textbf{Ablation study on the block size and in-block iteration steps. }
   }
   \label{fig:ablation}
   \vspace{-0.4cm}
\end{figure}

\begin{figure}[t]
    \centering

   \includegraphics[width=0.8 \linewidth]{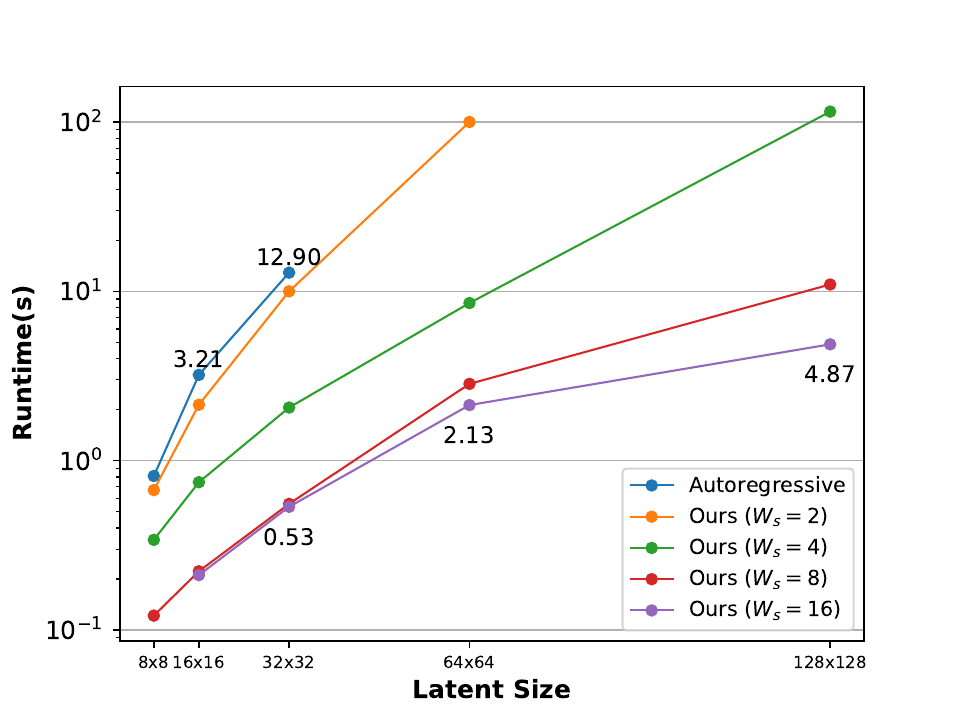}
\vspace{-0.3cm}
\caption{\textbf{Generation runtime comparison between ours with different block size $W_s$ and autoregressive method~\cite{esser2021taming}.}
   }
   \label{fig:runtime}
   \vspace{-0.3cm}
\end{figure}

\noindent \textbf{Block Size $W_s$.} We evaluate the generation performance with the block size of $2, 4, 8, 16$ in the left image of Fig.~\ref{fig:ablation}. When the block size is too small, \eg, $2 *2$, the model is prone to overfitting since there is only $2^4 = 16$ masked mode when learning local relationship within a block. It's worth noting that when block size is equal to latent size, i.e. $W_s=16$ here, our model will degrade into MaskGIT~\cite{chang2022maskgit}, that is, there is only one block during inference. Then all tokens will be predicted in a non-autoregressive manner. As we mentioned in Sec.~\ref{Training and Inference Strategy}, due to the large number of different masked patterns, the training phase cannot cover all the cases in inference stage. Besides, predicting many tokens simultaneously will lead to a more serious joint distribution problem, resulting in a bad FID score. The block size of $8$ obtains the best FID score since the suitable block size allows the model to learn both global semantics and local details relatively well. This comparison results shows the advantage of our sampling method over MaskGIT.
Besides, to study the effect of block size on generation speed, we compare autoregressive method, \ie, VQGAN and our model with different block sizes. As shown in Fig.~\ref{fig:runtime}, when $W_s=1$, our model performs as an autoregressive method. Ours with a larger block size runs faster, demonstrating the efficiency of the block-based design. 

\noindent \textbf{In-block Iteration Steps $T$.} We investigate the influence of the number of iterations within a block for image synthesis, as shown in the right image of Fig.~\ref{fig:ablation}. We assess the generated images as $T=4, 8, 12, 16, 24, 32$ on two models with different block sizes. When $T$ increases from $4$ to $8$, the FID gradually decreases since fewer steps means more tokens will be predicted simultaneously, which is hard to meet the independence assumption. However, more iteration steps do not lead to better quality because more iteration steps means more predictions based on known tokens, which narrows the sample probability distribution across the dataset, leading to less diversity. This observation is consistent with the findings in MaskGIT~\cite{chang2022maskgit}.


\section{Conclusion and Discussion}

This paper proposes Efficient-VQGAN, an efficient two-stage vector quantized model, for high-resolution image generation.
We make several improvements in both the first quantization and second generative modeling stage, contributing to higher computational efficiency and generation quality. A new image generation paradigm is developed that combines the masked autoencoding training and autoregressive inference, facilitating better generation quality. 
As for the limitations, we sample multiple tokens in parallel within each block, increasing the sampling speed at the cost of slightly reducing the quality of generated images.
Devising a better inference strategy to combine multi-grained attention can be interesting future work.

\blfootnote{\hspace{-0.2cm}This work is supported in part by the National Key R\&D Program of China(Grant No. 2022ZD0116403), National Natural Science Foundation of China (Grant No. 61721004 and Grant No. 62176255), the Strategic Priority Research Program of Chinese Academy of Sciences(Grant No. XDA27000000), and the Youth Innovation Promotion Association CAS.}

{\small
\bibliographystyle{ieee_fullname}
\bibliography{egbib}
}

\end{document}